\documentclass{article}

\usepackage{draftfigure} % For \placeholder command
\usepackage{booktabs, colortbl, multirow, xcolor}
\usepackage{amsmath,amssymb}
\usepackage[english]{babel}
\usepackage{algorithm}
\usepackage{algpseudocode}
\usepackage{float}      % provides the [H] placement specifier
\usepackage{graphicx}
\usepackage{subcaption}
\usepackage[section]{placeins} % keep floats inside their section
\usepackage{listings}
\lstdefinestyle{prompt}{
    basicstyle=\ttfamily\footnotesize,
    backgroundcolor=\color{gray!8},
    frame=single,
    rulecolor=\color{gray!55},
    framerule=0.4pt,
    breaklines=true,
    breakatwhitespace=true,
    columns=fullflexible,
    keepspaces=true,
    showstringspaces=false,
    captionpos=b,
    abovecaptionskip=4pt,
    aboveskip=8pt,
    belowskip=8pt,
    xleftmargin=4pt,
    xrightmargin=4pt,
}

\newtheorem{assumption}{Assumption}
\usepackage{xcolor}

\newif\ifcameraready
\camerareadytrue
\DeclareRobustCommand{\cameraready}[1]{\ifcameraready{\color{black}#1}\else#1\fi}

\usepackage[preprint]{corl_2026} % Uncomment for pre-prints (e.g., arxiv); This is like ``final'', but will remove the CORL footnote.

\newcommand{\Prob}{\mathbb{P}}                  % probability operator
\newcommand{\R}{\mathbb{R}}                     % real numbers
\newcommand{\Norm}{\mathcal{N}}                 % Gaussian / normal distribution

\newcommand{\Classes}{\mathcal{K}}              % semantic class set
\newcommand{\Hypset}{\mathcal{H}}               % hypothesis set

\newcommand{\mtgt}{\mathbf{m}^{\dagger}}        % target voxel indicator vector
\newcommand{\mprior}{\mathbf{m}^{-}}            % prior scene-graph map
\newcommand{\mvox}[1]{m_{#1}^{\dagger}}         % per-voxel target indicator

\newcommand{\Lhist}{L_{1:M}}                    % full utterance history
\newcommand{\Zhist}{Z_{1:T}}                    % full observation history

\newcommand{\bel}{b}                            % belief
\newcommand{\logodds}{\ell}                     % per-voxel log-odds

\newcommand{\hyp}{h}                            % grounded hypothesis
\newcommand{\hypweight}{w}                      % hypothesis weight
\newcommand{\refobj}{o}                         % candidate referent object

\newcommand{\gmean}{\mu}                        % predicted mean
\newcommand{\gcov}{\Sigma}                      % predicted covariance

\newcommand{\vlmap}{VL-Map}                     % system name
\newcommand{\lsm}{LSM}                          % language sensor model

\title{Language as a Sensor: Calibrated Spatial Belief Estimation in 3D Scenes from Natural Language
}

\author{
Aryan Naveen$^{1,2,*}$\hspace{1.2em}
Jason Xinyu Liu$^2$\hspace{1.2em}
Luca Carlone$^{1,\dagger}$\hspace{1.2em}
Andreea Bobu$^{2,\dagger}$\\[0.4em]
\\
$^1$ MIT Laboratory for Information \& Decision Systems\\
$^2$ MIT Computer Science \& Artificial Intelligence Laboratory\\
$^\dagger$ Equal advising
}

\begin{document}
\maketitle
\begin{figure}[H]
    \centering
    %\vspace{-9mm}
    \includegraphics[width=0.9\textwidth]{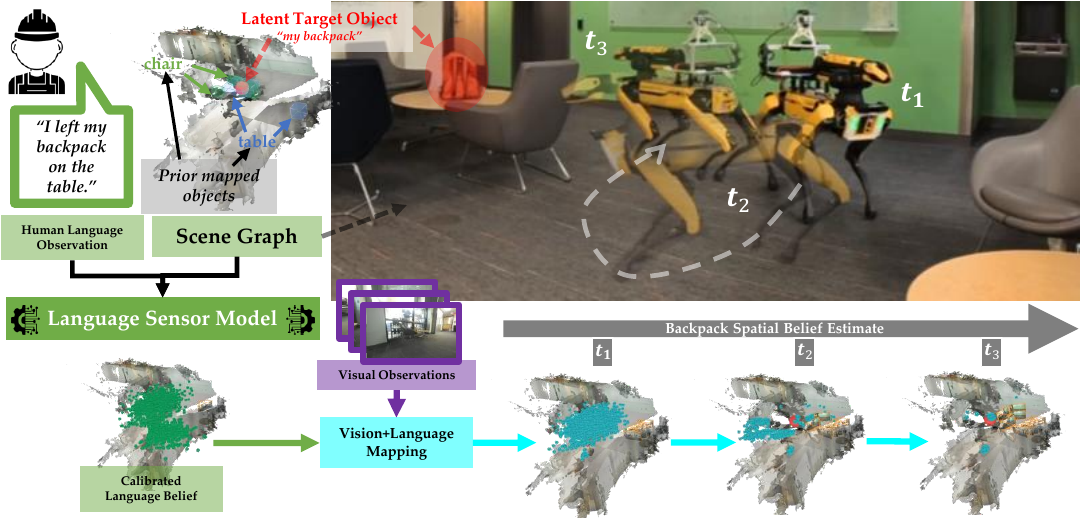}
    \caption{\textbf{Language as a complementary sensing modality.} A robot navigating a scene receives the utterance \emph{``I left my backpack on the table.''} The Language Sensor Model (\lsm{}) maps this utterance and the prior scene graph into a calibrated spatial distribution, which \vlmap{} fuses with streaming visual observations. By treating the utterance as a stochastic spatial observation, \vlmap{} initially places multimodal probabilistic mass over ambiguous table regions, which progressively concentrates on the true location as it fuses visual observations  .}
    \label{fig:motivating}
    %\vspace{-4mm}
\end{figure}
%===============================================================================
\begin{abstract}
Robots deployed in human-centric environments routinely receive natural-language descriptions of spatial information (``I left my backpack on the table'') that reference parts of the world beyond their perceptual field of view. Traditional metric-semantic mapping ignores this signal, while off-the-shelf multimodal models remain limited in 3D spatial reasoning and are not directly amenable to fusion with other sensor modalities. To convert language observations into a calibrated spatial distribution, we train a Language Sensor Model (\lsm{}) that maps each utterance and its scene-graph context to a multimodal distribution, with mixture weights encoding referential ambiguity (e.g., \textit{which table}) and component covariances encoding spatial uncertainty (e.g., \textit{where on the table the target lies}). We then introduce \vlmap{} (\emph{Vision-Language Metric-Semantic Mapping}), a probabilistic framework that treats these language predictions as stochastic observations and fuses them with onboard perception within a unified belief map. On the VLA-3D benchmark as well as on a real-world mobile robot, \lsm{} is the only language predictor whose covariance estimates remain within the calibrated regime; fused into \vlmap{}, it
leads to more accurate predictions of the target object location (${\sim}70\%$ more probability mass on the true target compared to the strongest foundation-model baseline). \\ \cameraready{\textbf{Project Website:} https://aryan-naveen.github.io/language-as-a-sensor/}
\end{abstract}
%\vspace{-4mm}

\section{Introduction}
\label{sec:intro}
\vspace{-3mm}

Humans possess the innate ability to reason not only about what they observe, but also about unobserved aspects of the world by leveraging prior knowledge, experience, and language. When given spatial information, such as \textit{``I left my backpack on the table,''} a person can ground a spatial belief in the environment's latent structure, even without direct visual confirmation \cite{peacock2019verbal, munnich2001spatial, hermer1999sources}.  For robots operating in human-centric environments, this same capability is essential: language often conveys spatial structure beyond the robot's field of view. Yet transforming ambiguous natural language into a calibrated spatial belief over world structure, as illustrated in Figure~\ref{fig:motivating}, remains largely unsolved.

Despite the prevalence of such linguistic cues, traditional metric-semantic mapping systems rely exclusively on onboard sensors to construct actionable representations of the environment \cite{rosinol2020kimera, hughes2022hydra, bowman2017probabilistic}. As a result, a robot told where its user's backpack is must still discover it from scratch, ignoring informative prior signals. The core challenge is that language is an inherently underspecified spatial sensing modality. Even a statement such as \textit{``the backpack is on the table''} admits multiple plausible interpretations: which table, what region of the surface, and how confidently the speaker knows the location. Humans resolve this ambiguity through prior knowledge about typical environments and object arrangements; robots instead require an explicit probabilistic representation that preserves and propagates uncertainty rather than collapsing language into an overconfident guess.

Recent 3D Vision-Language Models (3D-VLMs) bridge language and 3D geometry for retrospective reasoning over static scenes \cite{zhu20233d, hong20233d, chen2020scanrefer}, but produce deterministic groundings rather than calibrated spatial distributions, limiting their ability to fuse language with streaming observations or reason probabilistically over unobserved regions in a robot's evolving belief state. We address this gap with \textbf{\vlmap{}} (\emph{Vision-Language Metric-Semantic Mapping}), a framework that treats language as a spatial sensor measurement over latent scene structure. Language-derived likelihoods fuse with streaming visual evidence into an unified voxel-level belief representation. We instantiate \vlmap{} within Hydra \cite{hughes2022hydra}, enabling language-derived priors to be progressively refined as the robot explores.

This fusion is only as useful as the calibration of the language sensor itself:
overconfident language predictions can actively degrade the posterior, leaving
the robot worse off than if it had ignored language entirely. We therefore
develop a transformer-based \textbf{Language Sensor Model} (\lsm{}) that maps
each utterance and its constructed prior scene graph into a Gaussian mixture
whose weights encode \emph{referential ambiguity}---which entity the speaker
refers to---and whose component covariances encode \emph{spatial ambiguity}---where
on or near it the target lies. We validate \vlmap{} both in simulation on the
VLA-3D benchmark and on a real-world mobile robot: \lsm{} is the only language
predictor we evaluated that retains calibration across spatial relations and
ambiguity levels, and the only one whose integration into \vlmap{} improves
rather than degrades the fused posterior. Our contributions are summarized
as follows:
\begin{enumerate}
    \item \textbf{Language Sensor Model.} We develop a transformer-based \lsm{} that maps unstructured language into calibrated Gaussian-mixture spatial distributions representing both referential and spatial ambiguity. (Section~\ref{sec:lsm})

    \item \textbf{\vlmap{} framework.} We introduce a Bayesian formulation that treats natural language as a probabilistic sensing modality, directly updating a robot's spatial belief over latent scene structure using both language and onboard perception. (Section~\ref{sec:vlmap})

    \item \textbf{Empirical validation.} We show that \lsm{} is the only evaluated language predictor that
retains calibration; when fused into \vlmap{}, it assigns ${\sim}70\%$ more posterior probability mass to the true target than the strongest foundation-model grounding baseline. (Section~\ref{sec:experiments})
\end{enumerate}

%===============================================================================
%===============================================================================
\vspace{-4mm}
\section{Related Works}
\label{sec:relatedworks}
\vspace{-2mm}

\textbf{3D Vision-Language Foundation Models.} Recent 3D vision-language models align point clouds or scene-graph tokens with natural language for open-vocabulary 3D understanding, spanning transformer-based grounders (3D-VisTA~\cite{zhu20233d}, MVT~\cite{huang2022mvt}), generative multimodal architectures (3D-LLM~\cite{hong20233d}, 3D-LLaVA~\cite{deng20253dllava}), and scaled multi-task pretraining (SceneVerse~\cite{jia2024sceneverse}). These models are evaluated on visual-grounding (ReferIt3D~\cite{achlioptas2020referit3d}) and 3D question-answering (ScanRefer~\cite{chen2020scanrefer}) benchmarks that assume a fully observed scene and reward a single deterministic prediction per utterance. Two consequences make them unsuitable as a real-time language sensor: (i) they emit no calibrated uncertainty over their grounding, precluding Bayesian fusion with onboard observations, and (ii) they cannot reason over targets absent from the input scene---precisely the regime our problem formulation targets (Section~\ref{sec:problemform}). Our \lsm{} (Section~\ref{subsec:lss}) instead predicts an explicit Gaussian-mixture spatial distribution conditioned on a \emph{partial} scene graph, retaining open-vocabulary generality while exposing the probabilistic interface that \vlmap{}'s closed-loop belief requires.

\textbf{Language Model Uncertainty Quantification.} Two paradigms dominate uncertainty quantification for language models. \emph{Self-assessed} approaches prompt the model to verbalize or score its own confidence~\cite{kadavath2022langmodels, xiong2024llmuncertainty,peng2024plga}, but consistently exhibit miscalibration and systematic overconfidence---particularly on long-tail inputs where calibration matters most. \emph{Conformal prediction} provides distribution-free coverage guarantees~\cite{kumar2023conformal, mohri2024conformalfactuality,ren2023knowno} but becomes severely conservative when ported to continuous high-dimensional outputs such as 3D coordinates, where the conformal set degenerates into a near-uninformative region of space. Neither paradigm yields the calibrated continuous spatial distribution required to use language as a probabilistic sensor inside a framework like \vlmap{}; our \lsm{} (Section~\ref{subsec:lss}) is trained end-to-end with a probabilistic regression loss and empirically \cameraready{outperforms foundation model-based predictors---including post-hoc-calibrated ones---in both accuracy and calibration at every ambiguity level} (Section~\ref{subsec:grounding}).

\textbf{Language Guided World Understanding.} Prior work has explored language as a signal about world state for embodied agents~\cite{languageunderstandingfieldrobot,kim2024lingospacelanguageconditionedincrementalgrounding,nader2026quicklapquicklanguageactionpreference,peng2024LGA,peng2024algae, hwang2026maskedirl}. Walter et al.~\cite{languageunderstandingfieldrobot} similarly treat language as a sensor, but rely on hand-designed grammars and fixed symbolic representations that lack the open-vocabulary generality of modern VLMs and do not produce calibrated continuous spatial likelihoods. LINGO-Space~\cite{kim2024lingospacelanguageconditionedincrementalgrounding} incrementally grounds composite referring expressions, but assumes fully observed scenes and only localizes already-perceived objects. QuickLAP~\cite{nader2026quicklapquicklanguageactionpreference} demonstrates that language can serve as calibrated evidence in Bayesian inference, though over reward functions rather than spatial beliefs under partial observability. In contrast, \lsm{} (Section~\ref{subsec:lss}) predicts calibrated continuous spatial distributions from partial scene graphs, while \vlmap{} (Section~\ref{sec:vlmap}) recursively fuses those distributions with onboard perception through the belief update of Eq.~\ref{eq:belief}.

%===============================================================================
%===============================================================================
\vspace{-3.5mm}
\section{Problem Setup}
\label{sec:problemform}
\vspace{-3.5mm}

We consider a robot navigating an environment while tasked with localizing a target object that may lie outside its current
perceptual field. As it moves, the robot receives natural-language utterances
describing the target object location (e.g., \textit{``I left my backpack on the table''}) and
has access to a prior scene-graph map of the environment from an earlier
mapping pass, which does not contain the desired object. Our objective is to translate each utterance into a calibrated
probabilistic distribution over the target's location---one that
downstream inference can fuse with onboard sensing modalities.

\noindent\textbf{Target object.} We denote by $\mtgt$ the latent random variable
encoding the geometric state of a target object in the workspace. In our
running example, $\mtgt$ corresponds to the unobserved location of the
backpack referenced in the utterance.

\textbf{Language observations.} The robot receives a set of natural-language
utterances $\Lhist = \{L_j\}_{j=1}^M$ describing the target location. These utterances
are decoupled from the robot's timesteps: the human may provide a description
once, sporadically, or repeatedly throughout deployment, and our framework
absorbs each new utterance as it arrives.

\textbf{Prior scene graph.} The robot has access to a prior map $\mprior$ from an earlier mapping pass: a symbolic instance-level scene graph whose nodes represent previously mapped objects grouped hierarchically into regions~\cite{hughes2022hydra}. We adopt this representation because language refers to semantic entities such as tables, drawers, and rooms, yet the map is often incomplete, omitting small or occluded targets (e.g., backpacks or keys) even when their surrounding structure is present.

\textbf{Objective.} Given an utterance $L_j$ and the prior scene graph
$\mprior$, we seek a probabilistic distribution over the target's latent state: $\Prob\!\left(\mtgt \,\big|\, L_j,\, \mprior\right)$. This distribution must be calibrated and
fusible with other sensing modalities so that downstream recursive inference can
refine the belief over time. Section~\ref{sec:lsm} models this language
sensor; Section~\ref{sec:vlmap} shows how to fuse language priors with onboard perception into an
unified spatial belief.

%===============================================================================
\begin{figure}[t]
    \centering
    \includegraphics[width=\textwidth]{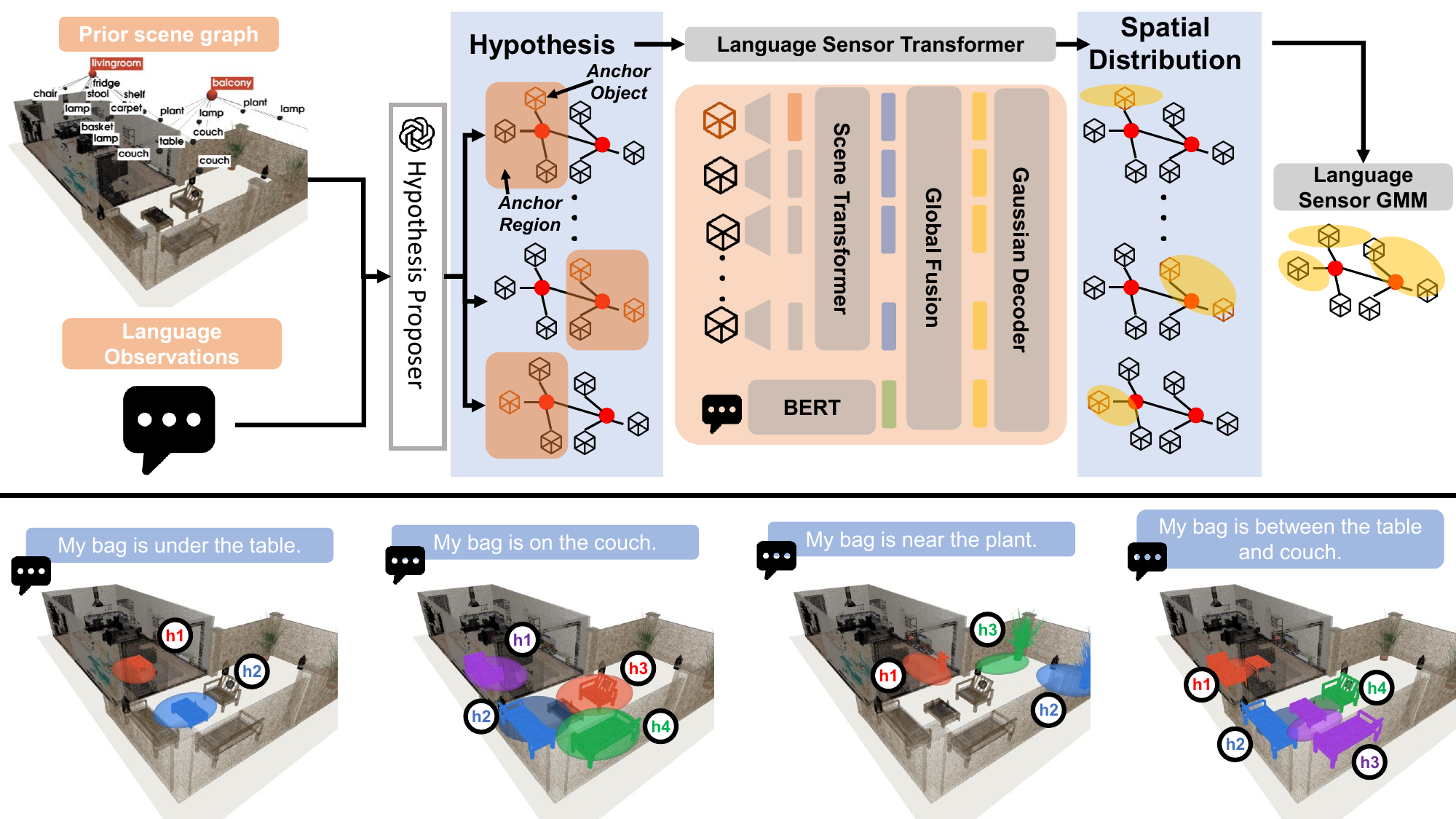}
    \caption{\textbf{The Language Sensor Model (LSM) instantiates Eq.~\ref{eq:language_mixture} as a learned Gaussian mixture over the target's 3D position.} \emph{Top:} the prior scene graph $\mprior$ and a language observation $L_j$ are passed to a hypothesis proposer, a Large Language Model, that enumerates anchor candidates (each scoped to an anchor object and region) and assigns each a confidence; for every hypothesis $\hyp_i$, the Language Sensor Transformer fuses object tokens (Scene Transformer) with the utterance encoding (BERT) and decodes a Gaussian, which combine into the language-sensor GMM. \emph{Bottom:} qualitative multimodal LSM outputs for four spatial relations on the same scene---\emph{under}, \emph{on}, \emph{near}, \emph{between}---with anchors colored by the mode they induce.}
    \label{fig:lsm_pipeline}
\end{figure}
%===============================================================================
\vspace{-2mm}
\section{Language Sensor Model}
\vspace{-3mm}
\label{sec:lsm}
\label{subsec:lss}

In this section, we model the language distribution $\Prob(\mtgt \mid L_j, \mprior)$---the probabilistic distribution over the target's latent state given a single utterance and the prior scene graph.

Consider a robot receiving the utterance \textit{``my backpack is on the table''} in an environment containing two tables. The core challenge is translating this utterance into a calibrated spatial probability distribution over the backpack's location. Two intertwined sources of uncertainty make this challenging: \emph{referential ambiguity}---which table the speaker means---and, conditioned on a table, \emph{spatial ambiguity}---where on that table the backpack lies. A faithful language sensor must represent both.

A language observation is often consistent with multiple scene descriptions. We capture this through a hypothesis space $\Hypset(L_j, \mprior) := \{\hyp_i = (\refobj_i^{(r)}, \rho_i)\}_{i=1}^{K}$, where each hypothesis $\hyp_i$ pairs a candidate referent object $\refobj_i^{(r)} \in \mprior$ with a grounded spatial relation $\rho_i$ (\emph{on}, \emph{in}, \emph{near}, \emph{between}, \dots) extracted from the utterance; for our running utterance, $\Hypset = \{(\text{\texttt{Table 1}}, \text{on}),\, (\text{\texttt{Table 2}}, \text{on})\}$ when $\mprior$ contains two table instances. We model the full language observation as a mixture over $\Hypset$ that factors the two sources of uncertainty identified above:
\begin{equation}
    \Prob\!\left(\mtgt \,\big|\, L_j, \mprior\right)
    \;=\;
    \sum_{\hyp_i \in \Hypset}
    \underbrace{\Prob\!\left(\mtgt \,\big|\, \hyp_i, \mprior\right)}_{\text{spatial grounding}}
    \;\cdot\;
    \underbrace{\Prob\!\left(\hyp_i \,\big|\, L_j, \mprior\right)}_{\text{referent disambiguation}}\enspace
    \label{eq:language_mixture}
\end{equation}

Note that we assume $\mtgt$ is conditionally independent of $L_j$ given $h_i$, due to the fact that $h_i$ encodes the information available within the initial language observation. Next, we describe a Language Sensor Model (\lsm{}) that implements the two factors in Eq. (\ref{eq:language_mixture}).

\textbf{Hypothesis Generation.} We instantiate the referent-disambiguation factor of Eq.~\ref{eq:language_mixture} with a large language model prompted with the utterance and a region-grouped JSON serialization of the scene graph $\mprior$ (Appendix~\ref{app:hyp_prompt}). The LLM returns a weighted set $\{(\hyp_i, \hypweight_i)\}_{i=1}^H$ of scene-graph entities satisfying the spatial relation, softmax-normalized so that $\hypweight_i \approx \Prob(\hyp_i \mid L_j, \mprior)$. \cameraready{The proposer also normalizes lexical paraphrases into grounded spatial
relations and object references. We analyze its precision and recall
across ambiguity levels and ablate the size of the requested hypothesis set in Appendix~\ref{app:hypothesis_proposer}.}

\textbf{Spatial Grounding. }Given a fixed referent, predicting where the target lies requires geometry-aware reasoning over the anchor region, not just language alone---the phrase \emph{``on the table''} grounds the backpack to a different spatial distribution depending on whether the table is small and central or large and pushed against a wall. We train a transformer that takes a hypothesis $\hyp_i$ together with its scene context $\mprior$ and emits a Gaussian $\Norm(\gmean_i,\gcov_i)$ over $\R^3$ representing the target's likely position.

The architecture proceeds in two sequential stages (Fig.~\ref{fig:lsm_pipeline}, top). A \emph{spatial transformer backbone} trained from scratch, following \cite{zhu20233d}, refines object tokens via geometry-aware self-attention prior to any language conditioning. A small MLP maps the anchor-centric centroids and sizes of each object pair to a learned pairwise bias on the attention logits, encoding the anchor region's relational layout into the object tokens. A \emph{global fusion backbone} then concatenates the text CLS token (encoded by a frozen pretrained language model) with the spatially refined object tokens and processes them with a standard multi-head transformer encoder, inducing bidirectional cross-modal reasoning between language and geometry. A FiLM-conditioned decoder head emits $(\gmean_i, \gcov_i)$ from the mean-pooled fused representation, with FiLM scaling on the local region extent so predictions transfer across environments of varying scale. We train the model with a combination of negative log-likelihood and direct Euclidean regression losses on the ground-truth target centroid; see Appendix~\ref{app:lsm_impl} for the full architecture dimensions and the exact training objective, and Appendix~\ref{app:ablations} for component ablations.

%===============================================================================
%===============================================================================
\vspace{-3mm}
\section{VL-Map: Fusing Language with Streaming Perception}
\label{sec:vlmap}
\label{subsec:formulation}
\vspace{-2mm}

Given the language-induced distribution $\Prob(\mtgt \mid L_j, \mprior)$, we now fuse it with streaming perceptual observations within a Bayesian framework that recursively updates a spatial belief over $\mtgt$.

\textbf{Voxel representation.} To inject language into approaches for mapping with traditional sensors, such as RGB-D cameras, we discretize the workspace into $N$ non-overlapping voxels indexed by $i \in [N]$. A binary indicator $\mtgt \in \{0,1\}^N$ encodes the state of the target object, with $\mvox{i} = 1$ iff the target occupies voxel $i$. This is the concrete instantiation of the latent target state introduced in Section~\ref{sec:problemform}; the indicator $\mtgt$ is the random variable over which the robot maintains a belief.

\textbf{Streaming observations.} As the robot moves, it visits a sequence of poses $\{x_t\}_{t=1}^T$ (each $x_t$ encoding position and orientation jointly), with $x_{1:t}$ denoting the pose history up to time $t$. It acquires RGB-D observations $\Zhist = \{Z_t\}_{t=1}^T$ annotated with inferred per-pixel semantic labels $y_{t,b} \in \Classes := \{0,1,\dots,K\}$, where $b$ indexes pixels.

\textbf{Recursive belief.} Given $\Zhist$, $\Lhist$, and $\mprior$, we recursively maintain the posterior belief
\begin{equation}
    \bel_t(\mtgt)
    \;:=\;
    \Prob\!\left(\mtgt \;\big|\; Z_{1:t},\, \Lhist,\, x_{1:t},\, \mprior\right)\enspace.
    \label{eq:belief}
\end{equation}
To make joint inference tractable, we extend the standard conditional independence assumption across observations, including both language and visual sources.

\begin{assumption}
\label{ass:cond_ind}
Conditioned on the map $\mtgt$, the language utterances $\Lhist$ and perceptual observations $\Zhist$ are conditionally independent:
\begin{equation}
\Prob(\Lhist, \Zhist \mid \mtgt) = \prod_{t=1}^T \Prob(Z_t \mid \mtgt) \cdot \prod_{j=1}^M\Prob(L_j \mid \mtgt)\enspace.
\end{equation}
\end{assumption}

Assumption~\ref{ass:cond_ind} allows the posterior belief to decompose into visual and linguistic observation models:
\begin{equation}
    \bel_t(\mtgt)
    \propto
    \underbrace{
    \Prob\!\left(
        Z_{1:t}
        \mid
        \mtgt,
        x_{1:t}
    \right)
    }_{\text{Visual Observation Model}}
    \;
    \underbrace{
    \prod_{j=1}^{M}
    \Prob\!\left(
        \mtgt
        \mid
        L_j,
        \mprior
    \right)
    }_{\text{Language Observation Model}}\enspace.
\label{eq:vlmap_belief_factorization}
\end{equation}
This formulation interprets language as a probabilistic sensing modality analogous to a conventional observation model: each utterance contributes a spatial likelihood over the target's geometric state, which combines with visual evidence from onboard perception (see Appendix~\ref{app:belief_derivation} for a derivation).

The LSM language factor plugs directly into any voxel-level mapper supporting additive log-odds updates. We integrate with Hydra~\cite{hughes2022hydra}, where visual and language log-odds accumulate additively per voxel, with the two pipelines running asynchronously: the visual update runs at sensor frame rate, while each new utterance triggers a single language update. The full per-voxel fusion equation and derivation are in Appendix~\ref{subsec:integration}.

\vspace{-3mm}
\section{Experiments}
\label{sec:experiments}
\vspace{-3mm}

We evaluate \lsm{} as a static language sensor and \vlmap{} as a closed-loop language--vision fusion system, on the VLA-3D simulation benchmark \cite{zhang2024vla3ddataset3dsemantic} and on a Boston Dynamics Spot, to answer two questions:
(i)~\emph{Static grounding:} Does \lsm{} produce more accurate and better-calibrated spatial distributions than foundation-model language groundings, and does this advantage hold as utterances become more ambiguous and spatial relations more underspecified?
(ii)~\emph{Closed-loop fusion:} When we plug language into the \vlmap{} framework and fuse with streaming visual observations, does \lsm{}'s use translate into more informative belief updates over time?

% -----------------------------------------------------------------------------
\vspace{-1mm}
\subsection{Experimental Setup}
\label{subsec:setup}
\vspace{-1mm}

\textbf{Simulation Dataset and Task.} 
We evaluate on the VLA-3D benchmark~\cite{zhang2024vla3ddataset3dsemantic}, which contains over 21,000 language-grounded object descriptions across thousands of indoor scenes from Matterport3D \cite{chang2017matterport3d}, 3RScan \cite{dai2017scannet}, and ARKitScenes \cite{baruch2021arkitscenes}. By removing the target node prior to grounding, we test the model's ability to infer latent objects across both \texttt{val\_seen} and \texttt{val\_unseen} splits (2058 utterances per split in Table~\ref{tab:unambig_grounding}; 686 \texttt{val\_seen} utterances per ambiguity level in Fig.~\ref{fig:ambiguity_vs}) to measure in-distribution generalization and cross-scene transfer. \cameraready{We train \lsm{}'s spatial predictor for 50 epochs, holding out entire environments for unseen-scene validation. Training includes object-dropout and pose-noise augmentation; implementation details are provided in Appendix~\ref{app:lsm_implementation}.} 
% Furthermore, we partition these utterances by their semantic ambiguity, defined as the number of valid anchor entities that match the spatial relation, to stress-test how models navigate uncertainty in multimodal interpretation.

\textbf{Baselines.} We compare \lsm{} against three foundation-model language sensors and a language-free \textbf{vision-only} lower bound. In the closed-loop evaluation, all methods share \vlmap{}'s Hydra-based visual integrator~\cite{hughes2022hydra}, isolating differences to the language observation model alone. \textbf{LLM-E2E} directly predicts the full Gaussian mixture from the utterance and serialized scene context, testing whether explicit decomposition is necessary. \textbf{Scaffolded-LLM} uses an LLM to enumerate hypotheses and regress per-hypothesis Gaussians, isolating the contribution of \lsm{}'s learned spatial predictor within the same decomposition. \textbf{Scaffolded-VLM} replaces the per-hypothesis predictor with a VLM conditioned on bird's-eye-view crops, evaluating whether frontier VLMs can exploit the same geometric grounding signal used by \lsm{} (implementation details and prompts are provided in Appendix~\ref{app:baseline_prompts}). \cameraready{To test whether calibration can be recovered without a learned geometric predictor, we additionally evaluate three post-hoc-calibrated variants of Scaffolded-LLM---rescaling its covariances by a global or per-axis factor fitted on a held-out calibration split, or supplying measured per-relation covariances as in-context examples---and \textbf{3D-ViSTA}~\cite{zhu20233d} fine-tuned with a GMM head on the same training dataset as \lsm{}'s spatial predictor.}

\textbf{Metrics.} For isolated spatial grounding, we evaluate accuracy and calibration using \textbf{RMSE}, \textbf{NLL}, and \textbf{ANEES}---where an ANEES of exactly $3$ indicates a perfectly calibrated $3$D Gaussian, while values greater than $3$ reflect overconfidence and values less than $3$ reflect underconfidence. For closed-loop evaluation, we measure online performance using \textbf{Information Gain (IG)} and the \textbf{mean target probability mass} accumulated over a rollout. Finally, we track \textbf{Success \%} (significant terminal target mass) and \textbf{Miss \%}  (terminal target mass is effectively 0) to quantify how often the model achieves correct, confident beliefs versus overconfident failures, with intermediate scores capturing cases of unresolved ambiguity.

% -----------------------------------------------------------------------------
\vspace{-3mm}
\subsection{Static Spatial Grounding Evaluation}
\vspace{-1mm}
\label{subsec:grounding}

\textbf{\lsm{} outperforms all baselines in both accuracy and uncertainty quantification.} To disentangle errors arising from referent disambiguation versus spatial grounding in Eq. \eqref{eq:language_mixture}, we first evaluate the conditional predictor $\Prob(\mtgt \mid \hyp_i, \mprior)$ in isolation. Concretely, each method is provided the ground-truth hypothesis $\hat{h}$, and we measure the resulting estimate $\Prob(\mtgt \mid \hat{h}, \mprior)$. Table~\ref{tab:unambig_grounding} shows \lsm{} attaining the lowest RMSE and NLL on both splits and the least overconfident covariances (ANEES $1.83$ on \texttt{val\_seen}, $3.42$ on \texttt{val\_unseen}), whereas Scaffolded-LLM reaches ANEES values of $72$ and $172$, $24$--$57\times$ the calibrated target. \cameraready{Post-hoc calibration narrows but does not close this gap: lacking geometric priors, rescaling Scaffolded-LLM's covariances brings its \texttt{val\_seen} ANEES to $4.2$ but transfers poorly to held-out scenes ($8.2$--$9.9$) and cannot correct its means (RMSE unchanged, $2.5\times$ \lsm{}'s). 3D-ViSTA, which couples referential and spatial reasoning in a single network, remains $2.2$--$2.5\times$ less accurate than \lsm{}.} 
\begin{table}[!htb]
    \centering
    \small
\caption{Grounding accuracy and calibration on unambiguous utterances using the ground-truth anchor proposer. Green highlights the best results per column (ANEES closest to $3$).}
    \label{tab:unambig_grounding}
    \resizebox{\textwidth}{!}{%
    \begin{tabular}{l ccc ccc}
        \toprule
                       & \multicolumn{3}{c}{val\_seen} & \multicolumn{3}{c}{val\_unseen} \\
        \cmidrule(lr){2-4} \cmidrule(lr){5-7}
        Method         & RMSE $\downarrow$ & NLL $\downarrow$ & ANEES $\to 3$
                       & RMSE $\downarrow$ & NLL $\downarrow$ & ANEES $\to 3$ \\
        \midrule
        Scaffolded-LLM
                       & \cameraready{$1.78 \pm 2.87$}
                       & \cameraready{$35.82 \pm 345.06$}
                       & \cameraready{$72.01 \pm 690.23$}
                       & \cameraready{$3.68 \pm 5.50$}
                       & \cameraready{$86.04 \pm 370.05$}
                       & \cameraready{$171.91 \pm 740.27$} \\
        \cameraready{\quad + global rescale}
                       & \cameraready{$1.78 \pm 2.87$}
                       & \cameraready{$6.17 \pm 19.92$}
                       & \cameraready{$4.16 \pm 39.88$}
                       & \cameraready{$3.68 \pm 5.50$}
                       & \cameraready{$9.33 \pm 21.33$}
                       & \cameraready{$9.93 \pm 42.77$} \\
        \cameraready{\quad + per-axis rescale}
                       & \cameraready{$1.78 \pm 2.87$}
                       & \cameraready{$5.94 \pm 16.06$}
                       & \cameraready{$4.18 \pm 32.18$}
                       & \cameraready{$3.68 \pm 5.50$}
                       & \cameraready{$8.23 \pm 15.92$}
                       & \cameraready{$8.23 \pm 31.93$} \\
        \cameraready{\quad + in-context}
                       & \cameraready{$1.77 \pm 2.77$}
                       & \cameraready{$7.94 \pm 38.52$}
                       & \cameraready{$10.90 \pm 77.09$}
                       & \cameraready{$3.45 \pm 5.59$}
                       & \cameraready{$23.75 \pm 89.07$}
                       & \cameraready{$42.28 \pm 178.35$} \\
        % Scaffolded-VLM was NOT re-run for the camera-ready (VLM stages skipped):
        % these are the original-submission values, from a different run than the
        % rows above. Re-run with SKIP_VLM=0 before final submission.
        Scaffolded-VLM & \cameraready{$2.27 \pm 1.69$}  & \cameraready{$25.11 \pm 49.40$}  & \cameraready{$50.08 \pm 99.06$}
                       & \cameraready{$2.79 \pm 3.86$}  & \cameraready{$39.92 \pm 164.04$} & \cameraready{$79.57 \pm 328.03$} \\
        \cameraready{3D-ViSTA (fine-tuned)}
                       & \cameraready{$1.55 \pm 1.64$}
                       & \cameraready{$2.33 \pm 2.93$}
                       & \cameraready{$4.69 \pm 10.66$}
                       & \cameraready{$3.66 \pm 4.43$}
                       & \cameraready{$4.59 \pm 4.05$}
                       & \cameraready{$9.08 \pm 22.27$} \\
        \midrule
        \rowcolor{gray!8}
        \lsm{} (ours)
                       & \cellcolor{green!25}\cameraready{$0.71 \pm 0.69$}
                       & \cellcolor{green!25}\cameraready{$0.94 \pm 1.67$}
                       & \cellcolor{green!25}\cameraready{$1.83 \pm 1.99$}
                       & \cellcolor{green!25}\cameraready{$1.46 \pm 1.68$}
                       & \cellcolor{green!25}\cameraready{$4.17 \pm 8.71$}
                       & \cellcolor{green!25}\cameraready{$3.42 \pm 16.71$} \\
        \bottomrule
    \end{tabular}}
\end{table}

% \textbf{Foundation-model failure is concentrated on spatially-ambiguous relations.} Disaggregating by spatial relation (Appendix~\ref{app:per_relation}, Fig.~\ref{fig:unambiguous_performance}) reveals that the collapse is not uniform: the foundation-model baselines fail on uncertainty quantification for relations that carry high amounts of intrinsic spatial ambiguity---most prominently \emph{near}, \emph{between}, and \emph{above}---which are precisely the cases where the utterance genuinely underdetermines a valid region and a calibrated covariance has the most to do. Their per-hypothesis Gaussians inherit overconfident covariances regardless of which relation triggered the hypothesis, so the model sounds equally certain about \emph{``on the table''} (geometrically tight) and \emph{``near the sofa''} (geometrically loose). \lsm{} clusters near the calibrated optimum across all six relations, because the spatial transformer learns relation-conditioned covariance shapes rather than inheriting a generic prior.

\begin{figure}[!htb]
    \centering
    % Each panel is authored at 0.495\textwidth, the widest pair that still sits
    % on one line, so it is placed 1:1 with no rescaling of its type.
    \begin{subfigure}[t]{0.45\textwidth}
        \centering
        \includegraphics[width=\linewidth]{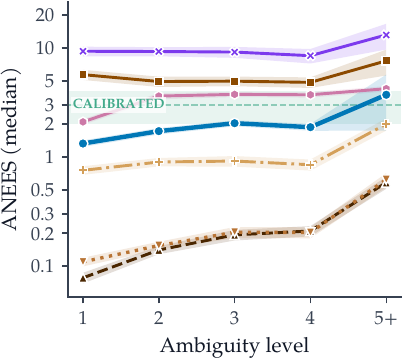}
        \caption{ANEES vs.\ ambiguity (calibration)}
        \label{fig:ambiguity_vs_anees}
    \end{subfigure}\hfill
    \begin{subfigure}[t]{0.45\textwidth}
        \centering
        \includegraphics[width=\linewidth]{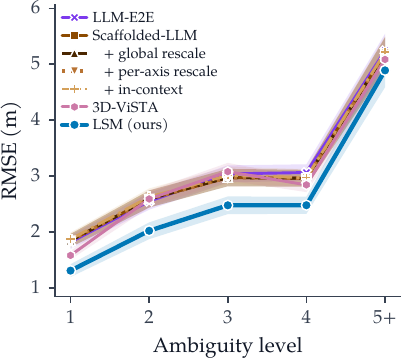}
        \caption{RMSE vs.\ ambiguity (accuracy)}
        \label{fig:ambiguity_vs_rmse}
    \end{subfigure}    
\caption{\cameraready{Model calibration (ANEES) and accuracy (RMSE) across ambiguity levels. The dashed line and shading indicate the ideal calibration band ($2-4$). \lsm{} achieves the lowest RMSE at every level, while its median ANEES never exceeds the band’s upper bound.}}
    \label{fig:ambiguity_vs}
    \vspace{-4mm}
\end{figure}

\textbf{Calibration matters more as ambiguity grows.} Figure ~\ref{fig:ambiguity_vs} sweeps ambiguity levels, where we observe that \lsm{}'s ANEES stays near 3 universally, while the foundation-model baselines’ already-overconfident per-hypothesis covariances compound across mixture components, driving
ANEES sharply away from the calibrated band and RMSE upward. \cameraready{Introducing post-hoc rescaling overshoots into underconfidence ($0.1$--$0.6$) because a single scale cannot correct heavy-tailed per-query errors, and 3D-ViSTA stays near the band but is $0.2$--$0.6$\,m less accurate at every level.} A secondary observation isolates the contribution of the referential-spatial decomposition: Scaffolded-LLM is consistently better calibrated than LLM-E2E on the multimodal cases despite sharing the LLM backbone, indicating that the Eq.~\ref{eq:language_mixture} split of hypothesis enumeration from per-hypothesis spatial regression is doing real work.

% -----------------------------------------------------------------------------
\subsection{Closed-Loop Belief Estimation}
\label{subsec:timevarying}

\begin{figure}[!htb]
    \centering
    \begin{subfigure}[t]{0.495\textwidth}
        \centering
        \includegraphics[width=\linewidth]{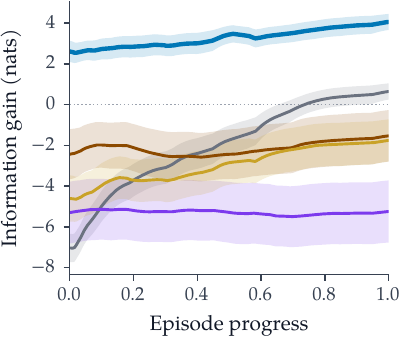}
        \caption{Information gain at the target surface (nats).}
        \label{fig:info_gain_time}
    \end{subfigure}\hfill
    \begin{subfigure}[t]{0.495\textwidth}
        \centering
        \includegraphics[width=\linewidth]{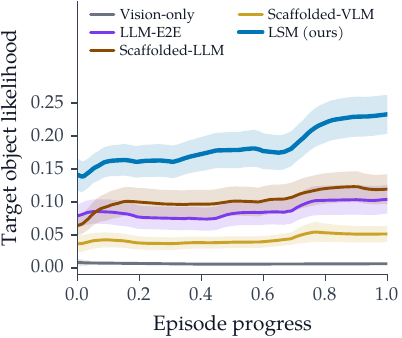}
        \caption{Mean target-object probability mass.}
        \label{fig:target_object_ll_time}
    \end{subfigure}
\caption{\cameraready{Belief evolution under language–vision fusion, measured by information gain and target-object probability over normalized episode progress. Only \lsm{} improves on the uninformed prior throughout the rollout; all foundation-model sensors degrade it.}}
    \label{fig:target_object_time_varying}
\end{figure}

\textbf{Bayesian fusion requires calibrated language distributions.} \cameraready{We plug each language predictor into \vlmap{} and fuse its predictions
with the same prerecorded visual observations, collected under a shared
random exploration policy. This holds the trajectory and visual evidence
fixed while varying only the language observation model.} Figure~\ref{fig:target_object_time_varying} traces information gain at the target location and mean target-object probability mass over normalized trajectory progress. Information gain measures improvement relative to a uniform spatial prior: positive values indicate that the fused belief assigns more likelihood to the true target location than assuming no prior knowledge, whereas negative values expose overconfident failure modes that suppress probability mass at the true target below the uninformed baseline. Under this metric, \lsm{} maintains positive information gain throughout the entire rollout (mean $+3.77$ nats), concentrating $22.3\%$ of the closed-loop belief mass on the true target voxel---$71\%$ more than the strongest foundation-model baseline (Scaffolded-LLM at $13.0\%$). Crucially, this advantage persists even when the target is never directly observed during the rollout ($n{=}10$): \lsm{} remains the only method with positive mean information gain ($+1.65$ nats) and succeeds in $70\%$ of episodes versus at most $40\%$ for any baseline. This result demonstrates that a calibrated language prior can maintain usable probability mass over unobserved regions, whereas overconfident language predictions actively degrade recursive state estimation relative to ignoring language entirely.

\textbf{Overconfidence causes catastrophic misplaced likelihood, not gradual drift.} The mechanism behind the foundation-model degradation is exactly the static-time pathology we exposed in Section~\ref{subsec:grounding}: an overconfident Gaussian commits the Bayes update to a confidently \emph{wrong} region, and once the posterior leaves the true support the filter cannot recover. In the observed regime of Table~\ref{tab:timevarying}, every foundation-model baseline ends below the vision-only baseline of $-0.60$ nats (LLM-E2E $-6.22$, Scaffolded-LLM $-1.85$, Scaffolded-VLM $-2.26$); the system would have been strictly better off ignoring language entirely. The failure shape is tail-dominated: foundation-model baselines miss $38.6\%$--$52.3\%$ of episodes (terminal mass below the uniform prior), versus $4.5\%$ for \lsm{}. This reduction in failures stems from \lsm{}'s ability to accurately calibrate the uncertainty associated with language observations, preventing overconfident estimates from poisoning the posterior.

% \textbf{A calibrated language prior extends the belief beyond the perceptual field.} The unobserved regime of Table~\ref{tab:timevarying} ($n{=}10$, target never directly observed during the rollout) is the setting that most directly tests \vlmap{}'s central premise: language is the \emph{only} available signal about the target, so the language sensor's calibration determines whether any usable belief exists at all. \lsm{} is the only method with positive mean information gain ($+1.65$ nats), concentrates $29.3\%$ of belief mass on the true target voxel ($1.9$--$3.7\times$ the foundation-model baselines), and succeeds in $70\%$ of episodes versus $0\%$ for vision-only and at most $40\%$ for any foundation-model baseline. In other words, a well-calibrated language prior places usable mass where the target actually is even without a direct visual observation, while an uncalibrated one cannot convert the same utterance into a usable spatial signal regardless of how confidently the language model phrases it.

\begin{table}[!htb]
    \centering
    \small
    \caption{Time-varying belief metrics across the rollout, split by whether the target was directly observed by vision during the trajectory. The \emph{Unobserved} regime isolates cases in which the robot never directly observes the target.}
    \label{tab:timevarying}
    \resizebox{\textwidth}{!}{%
    \renewcommand{\arraystretch}{0.92}
    \begin{tabular}{l ccc cc}
        \toprule
        Method            & Mean IG (nats) $\uparrow$
                          & Mean prob. mass $\uparrow$
                          & Mean argmax dist (m) $\downarrow$
                          & Success \% $\uparrow$
                          & Miss \% $\downarrow$ \\
        \midrule
        \multicolumn{6}{l}{\textit{Target observed by vision} ($n=44$)} \\
        \midrule
        Vision-only       & $-0.60 \pm 0.44$ & $0.006 \pm 0.001$ & $5.52$ & $0.0$  & $79.5$ \\
        LLM-E2E           & $-6.22 \pm 1.70$ & $0.095 \pm 0.024$ & $2.35$ & $22.7$ & $52.3$ \\
        Scaffolded-LLM    & $-1.85 \pm 1.35$ & $0.130 \pm 0.027$ & $2.25$ & $36.4$ & $38.6$ \\
        Scaffolded-VLM    & $-2.26 \pm 1.06$ & $0.046 \pm 0.010$ & $2.45$ & $9.1$  & $50.0$ \\
        \rowcolor{gray!8}
        \lsm{} (ours)     & \cellcolor{green!25}$3.77 \pm 0.33$
                          & \cellcolor{green!25}$0.223 \pm 0.030$
                          & \cellcolor{green!25}$1.41$
                          & \cellcolor{green!25}$52.3$
                          & \cellcolor{green!25}$4.5$ \\
        \midrule
        \multicolumn{6}{l}{\textit{Target never directly observed} ($n=10$)} \\
        \midrule
        Vision-only       & $-5.20 \pm 1.67$ & $0.007 \pm 0.003$ & $5.47$ & $0.0$  & $80.0$ \\
        LLM-E2E           & $-1.15 \pm 2.20$ & $0.155 \pm 0.054$ & $1.97$ & $40.0$ & $40.0$ \\
        Scaffolded-LLM    & $-2.93 \pm 2.30$ & $0.079 \pm 0.038$ & $3.27$ & $20.0$ & $40.0$ \\
        Scaffolded-VLM    & $-4.89 \pm 2.71$ & $0.083 \pm 0.054$ & $2.36$ & $10.0$ & $60.0$ \\
        \rowcolor{gray!8}
        \lsm{} (ours)     & \cellcolor{green!25}$1.65 \pm 1.38$
                          & \cellcolor{green!25}$0.293 \pm 0.086$
                          & \cellcolor{green!25}$1.75$
                          & \cellcolor{green!25}$70.0$
                          & \cellcolor{green!25}$30.0$ \\
        \bottomrule
    \end{tabular}}
\end{table}

\textbf{Hardware.} Across three closed-loop episodes spanning unique real-world environments on a Boston Dynamics Spot platform, we observe trends consistent with the simulation results. \cameraready{The prior scene graph is obtained from an actual earlier mapping pass, rather than ground truth; the deployment protocol is detailed in Appendix~\ref{app:deployment_protocol}.} Fusing the \lsm{} language channel with the streaming Hydra reconstruction attains a mean information gain at the target surface of $+4.34$ nats, and concentrates a mean of $8.7\%$ of the terminal posterior on the target surface, consistent with the calibrated-prior behavior observed in simulation and demonstrating that the scene-graph representation handles the sim-to-real gap. Figure~\ref{fig:motivating} makes this concrete on one of these scenes: \vlmap{} places probability mass on the true backpack location well before the robot directly observes it.

\vspace{-2mm}
\section{Conclusion}
\vspace{-2mm}
\label{sec:conclusion}

\subsection{Discussion}
We introduced \lsm{}, a learned language sensor model that converts language utterances and scene context into calibrated Gaussian-mixture spatial distributions through a referent-then-position decomposition. Building on this representation, we presented \vlmap{}, a Bayesian metric-semantic mapping framework that treats language as a probabilistic spatial sensor and fuses language-derived likelihoods with streaming visual observations in a unified voxel-level belief. Across simulation and real-world experiments, \lsm{} was the \cameraready{most accurate language model at every ambiguity level, including against post-hoc-calibrated and fine-tuned baselines}, and the only configuration whose integration into \vlmap{} consistently improved the robot's posterior rather than degrading it through overconfident grounding errors. Together, these results show that language can function as a reliable sensing modality for robotic spatial reasoning when uncertainty is modeled explicitly and fused probabilistically.

% -----------------------------------------------------------------------------
\subsection{Limitations and Future Work}
\label{subsec:limitations}

Our current \lsm{} implementation assumes that anchor objects referenced in an utterance already exist in the prior map $\mprior$, preventing grounding when no valid anchor hypothesis is available. \cameraready{Predictions also depend on the scene graph's geometric accuracy.
Sensitivity analysis of our augmentation-trained model shows tolerance
to 60\% object dropout when the true anchor is retained and nearly
unchanged NLL at a pose-noise level of $\sigma=0.1\,\mathrm{m}$;
larger pose errors degrade performance
(Appendix~\ref{app:scene_graph_robustness}).} In addition, the predicted target distribution is restricted to the mapped spatial extent of $\mprior$, assigning no probability mass to locations outside the explored region. Future work could address these limitations by detecting persistent disagreement between language-derived priors and streaming visual evidence, enabling the robot to reinterpret the utterance or request clarification from the human. \cameraready{While we focus on object localization in this work, the underlying insight suggests broader opportunities to leverage human expert priors as complementary sensor signals for tasks in human-centric environments.}

%===============================================================================

%===============================================================================

% \clearpage
% The acknowledgments are automatically included only in the final and preprint versions of the paper.
\acknowledgments{This work was supported by the ARL
DCIST program and MIT Energy Initiative (MITEI) Seed Award.}

%===============================================================================

\bibliography{example}  % .bib

\appendix
%===============================================================================
% Supporting figures and architecture details referenced from the main text.
% Include after \appendix in main.tex.
%===============================================================================
\newpage
\section{Integration with Metric-Semantic Mapping}
\label{subsec:integration}
\label{subsec:slam}
\label{subsec:visual_mapping}

The LSM produces a language factor $\Prob(\mtgt \mid L_j, \mprior)$ that any voxel-level semantic mapper supporting additive log-odds updates can consume. We demonstrate this by plugging into Hydra's real-time hierarchical scene-graph SLAM framework~\cite{hughes2022hydra}, whose semantic integrator already maintains a per-voxel log-odds belief $\logodds_i^{\mathrm{vis}}$ from the streaming RGB-D observations and per-pixel labels $y_{t,b}$ defined in Section~\ref{sec:vlmap}. By construction this visual likelihood is informative only at voxels the robot has actually observed; the LSM contributes a parallel likelihood that extends the belief beyond the robot's perceptual field.

Under Assumption~\ref{ass:cond_ind}, the visual and language likelihoods enter the per-voxel posterior $\logodds_i = \log \bel(\mvox{i} = 1)$ additively in log-space:
\begin{equation}
    \logodds_i^t
    \;=\;
    \underbrace{
        \sum_{\tau=1}^{t}
        \log p(y_{i,\tau} \mid \mvox{i} \!=\! 1)
    }_{\logodds_i^{\mathrm{vis}} \;\text{(Hydra integrator)}}
    \;+\;
    \underbrace{
        \sum_{j=1}^{M}
        \log p_{\mathrm{lang}}(\mvox{i} \mid L_j, \mprior)
    }_{\logodds_i^{\mathrm{lang}} \;\text{(LSM)}}\enspace,
    \label{eq:log_odds_fusion}
\end{equation}
Here $y_{i,\tau}$ is the semantic label assigned to voxel $i$ at time $\tau$, obtained by projecting the per-pixel observations $y_{\tau,b}$ onto voxel centroids via ray casting. We evaluate $p_{\mathrm{lang}}$ by querying the GMM of Eq.~\ref{eq:language_mixture} at the centroid of voxel $i$ and recover the posterior probability as $\bel_t(\mvox{i} \!=\! 1) = \sigma(\logodds_i^t)$. The two pipelines operate asynchronously: the visual update runs at sensor frame rate, while each new utterance triggers a language update.

\cameraready{\subsection{Deployment and Evaluation Protocol}
\label{app:deployment_protocol}

We first map the environment to construct a prior scene graph, then
place the target object and provide an utterance describing its
location relative to a previously mapped anchor. We subsequently
remap the environment and evaluate the evolving belief over the
target location. In the real-robot experiments, the prior scene graph
comes from an actual earlier mapping pass rather than ground truth.

All methods use the same prerecorded exploration sequence, holding
the trajectory and visual observations fixed while varying only the
language observation model. The vision-only baseline uses Hydra's
standard voxel integrator without language conditioning.}

\section{Derivation of the \vlmap{} Belief Factorization}
\label{app:belief_derivation}

We derive Eq.~\ref{eq:vlmap_belief_factorization} from the joint distribution over all variables, given Assumption~\ref{ass:cond_ind}.

\paragraph{Step 1: Proportionality.}
Since $\Prob(\Lhist, \Zhist, \mprior)$ is independent of $\mtgt$, it is a normalizing constant:
\begin{equation}
    \Prob\!\left(\mtgt \mid \Lhist, \Zhist, \mprior\right)
    = \frac{\Prob\!\left(\mtgt, \Lhist, \Zhist, \mprior\right)}{\Prob\!\left(\Lhist, \Zhist, \mprior\right)}
    \propto \Prob\!\left(\mtgt, \Lhist, \Zhist, \mprior\right)
\end{equation}

\paragraph{Step 2: Chain rule and conditional independence.}
Expanding the joint with the chain rule and treating $\mprior$ as fixed observed context:
\begin{align}
    \Prob\!\left(\mtgt, \Lhist, \Zhist, \mprior\right)
    &= \Prob\!\left(\Lhist, \Zhist \mid \mprior, \mtgt\right) \cdot \Prob\!\left(\mtgt \mid \mprior\right) \cdot \Prob\!\left(\mprior\right)
\end{align}
$\Prob(\mprior)$ does not depend on $\mtgt$ and is absorbed into the proportionality constant. Applying Assumption~\ref{ass:cond_ind} to factorize the joint observation term and expanding the language product:
\begin{align}
    \Prob\!\left(\mtgt, \Lhist, \Zhist, \mprior\right)
    &\propto \Prob\!\left(\Zhist \mid \mtgt, x_{1:T}, \mprior\right) \cdot \left[\prod_{j=1}^{M} \Prob\!\left(L_j \mid \mprior, \mtgt\right)\right] \cdot \Prob\!\left(\mtgt \mid \mprior\right)
\end{align}

\paragraph{Step 3: Rewriting each language likelihood via Bayes' rule.}
For each utterance $L_j$, Bayes' rule gives:
\begin{equation}
    \Prob\!\left(L_j \mid \mprior, \mtgt\right)
    = \frac{\Prob\!\left(\mtgt \mid \mprior, L_j\right) \cdot \Prob\!\left(L_j \mid \mprior\right)}{\Prob\!\left(\mtgt \mid \mprior\right)}
\end{equation}
Taking the product over all $M$ utterances and absorbing the $\mtgt$-independent factors $\Prob(L_j \mid \mprior)$ into the proportionality constant:
\begin{align}
    \Prob\!\left(\mtgt \mid \Lhist, \Zhist, \mprior\right)
    &\propto \Prob\!\left(\Zhist \mid \mtgt, x_{1:T}, \mprior\right) \cdot \frac{\displaystyle\prod_{j=1}^{M} \Prob\!\left(\mtgt \mid L_j, \mprior\right)}{\Prob\!\left(\mtgt \mid \mprior\right)^{M-1}}
    \label{eq:full_belief_derivation}
\end{align}

\paragraph{Prior correction.} We assume an uninformative uniform spatial prior over the target conditioned on the scene graph, therefore the denominator $\Prob(\mtgt \mid \mprior)^{M-1}$ is constant in $\mtgt$ and absorbed into the proportionality, recovering Eq.~\ref{eq:vlmap_belief_factorization} directly.
\section{Language Sensor Model Ablations}
\label{app:ablations}
\subsection{Model Architecture Ablation}
After training for $50$ epochs on the \texttt{val\_seen} split, we clearly observe that removing any of the components of \lsm{} causes degradation in performance.

\begin{figure}[!htb]
    \centering
    \includegraphics[width=0.9\textwidth]{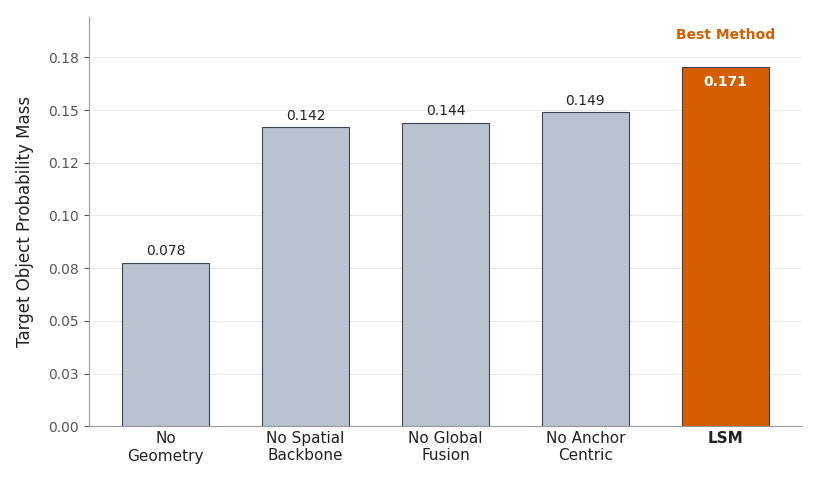}
    \caption{\textbf{\lsm{} component ablations on \texttt{val\_seen} after $50$ epochs.} Removing any single component degrades performance relative to the full LSM.}
    \label{fig:ablations}
\end{figure}

\cameraready{\subsection{Sensitivity to Scene-Graph Errors}
\label{app:scene_graph_robustness}

Since \lsm{} consumes structured scene-graph object tokens rather than
raw reconstruction geometry, we evaluate its sensitivity to object
dropout and pose-estimation noise. The evaluated model is trained with
both forms of augmentation.

\paragraph{Object dropout.}
\lsm{} remains robust at 60\% object dropout when the true referenced
anchor is guaranteed to remain present. This evaluates tolerance to
missing scene context, not the missing-anchor failure discussed in
the limitations.

\paragraph{Pose noise.}
Table~\ref{tab:scene_graph_pose_noise} reports NLL under increasing
object pose noise. NLL changes only slightly from $0.94$ to $0.96$
at $\sigma=0.1\,\mathrm{m}$, but increases to $1.65$ at
$0.5\,\mathrm{m}$ and $4.11$ at $1\,\mathrm{m}$.
These results indicate tolerance to small pose errors, while larger
perturbations degrade the predicted spatial distributions.

\begin{table}[!htb]
    \centering
    \small
    \caption{Sensitivity to object pose noise, measured by NLL.}
    \label{tab:scene_graph_pose_noise}
    \begin{tabular}{cc}
        \toprule
        Pose-noise level $\sigma$ (m) & NLL $\downarrow$ \\
        \midrule
        0   & 0.94  \\
        0.1 & 0.96  \\
        0.5 & 1.65  \\
        1   & 4.11  \\
        5   & 12.24 \\
        \bottomrule
    \end{tabular}
\end{table}}

\section{Hypothesis Proposer: Analysis and Prompts}
\label{app:hypothesis_proposer}

\subsection{Proposer Analysis}

\cameraready{We evaluate the hypothesis proposer, instantiated with GPT-5.2,
independently of the spatial predictor. The proposer maps lexical
paraphrases to normalized spatial relations and object references.
Ambiguity denotes the number of scene-graph hypotheses consistent
with an utterance.

Recall is particularly important: an omitted hypothesis is not passed
to the spatial predictor and therefore cannot contribute its own
Gaussian component. We evaluate precision and recall across ambiguity
levels and compare fixed hypothesis counts with dynamic selection.

% Insert the existing proposer precision--recall figure here.
% Suggested caption:
% Hypothesis-proposer precision and recall across ambiguity levels.

Table~\ref{tab:proposer_hypothesis_count} shows that dynamic selection
achieves the highest recall ($0.801$), while attaining nearly twice
the precision of the fixed-$k=5$ proposer ($0.418$ versus $0.211$).

\begin{table}[!htb]
    \centering
    \small
    \caption{Hypothesis-proposer precision and recall for fixed and
    dynamic hypothesis counts.}
    \label{tab:proposer_hypothesis_count}
    \begin{tabular}{lcc}
        \toprule
        Proposer & Precision $\uparrow$ & Recall $\uparrow$ \\
        \midrule
        $k=1$   & 0.538 & 0.538 \\
        $k=2$   & 0.413 & 0.698 \\
        $k=3$   & 0.320 & 0.768 \\
        $k=4$   & 0.259 & 0.782 \\
        $k=5$   & 0.211 & 0.786 \\
        \midrule
        Dynamic & 0.418 & \textbf{0.801} \\
        \bottomrule
    \end{tabular}
\end{table}}

\subsection{Prompts}
\label{app:hyp_prompt}

The system prompt used for the hypothesis-proposer LLM call in Section~\ref{sec:lsm}:

\begin{lstlisting}[style=prompt]
You are a spatial reasoning module.

You are given:

A scene graph describing a region and its objects.

A natural language utterance introducing a new object.

Your task is to generate K grounded spatial hypotheses that explain how the new object could relate to existing objects in the scene.

A hypothesis must include:

utterance: A fully specified sentence of the form
"there is a <target object> <relation> <existing object(s)>". Infer the target object from the provided utterance.

referred_object_id: A list of object ids the relation is grounded to. Use exactly one id for relations like "near", "on", "against", "above", "below". Use exactly two ids for "between" (both target objects, in any order).

region_id: The region id (integer) of the referred object(s). Same region for all referred objects in that hypothesis.

confidence: A number between 0 and 1 representing plausibility given the scene.

Only use objects that exist in the scene graph.
Do NOT invent objects.
Output MUST be valid JSON.
Return exactly K hypotheses.
\end{lstlisting}

The accompanying user template:

\begin{lstlisting}[style=prompt]
Scene Graph:
{scene_json}

Utterance:
"{utterance}"

Number of hypotheses (K):
as many as appropriate (at least 1, up to 5). Generate the minimum number of hypotheses necessary to disambiguate the utterance.

Instructions:
- Generate K diverse and plausible spatial hypotheses.
- Use only relations such as: near, between, above, below, in, or on. Avoid all other relations.
- referred_object_id must always be a list: one object id for single-object relations (e.g. "near the desk"), exactly two object ids for "between" (e.g. "between the desk and the wall" -> ["id1", "id2"]).
- Use only objects present in the scene graph.
- Apply a relational sparsity / non-redundancy prior: among multiple valid anchors, prefer objects that do not already participate in the same relation with an object of the target type.
- Confidence should reflect spatial plausibility and typical region layout priors.
- Consider different rooms and object arrangements to generate diverse hypotheses.
- Output format (referred_object_id is always a list; region_id is a single integer):

{
  "hypotheses": [
    {
      "utterance": "...",
      "referred_object_id": ["<object_id>"],
      "region_id": 0,
      "confidence": 0.8
    }
  ]
}
\end{lstlisting}

One of the three in-context demonstrations sent with every call, showing the scene-graph serialization format and the expected JSON response:

\begin{lstlisting}[style=prompt]
Utterance: "there is a mouse on the desk"

Scene Graph:
{
  "Office": {
    "objects": {
      "0": {"semantics": "desk",     "raw_label": "wooden desk",     "center": [0.0, -1.0, 0.4], "volume": 2.5,   "object_id": "0"},
      "1": {"semantics": "monitor",  "raw_label": "computer monitor","center": [0.1, -1.0, 0.9], "volume": 0.04,  "object_id": "1"},
      "2": {"semantics": "keyboard", "raw_label": "keyboard",        "center": [0.1, -0.85,0.7], "volume": 0.003, "object_id": "2"},
      "3": {"semantics": "chair",    "raw_label": "office chair",    "center": [0.8, -0.6, 0.5], "volume": 0.3,   "object_id": "3"},
      "4": {"semantics": "wall",     "raw_label": "wall",            "center": [0.0, -2.5, 1.5], "volume": 6.0,   "object_id": "4"},
      "5": {"semantics": "chair",    "raw_label": "folding chair",   "center": [1.8, -1.6, 0.5], "volume": 0.3,   "object_id": "5"}
    },
    "region_id": 0
  },
  "Lobby": {
    "objects": {
      "6": {"semantics": "sofa",  "raw_label": "sofa",         "center": [0.0, -1.0, 0.5], "volume": 1.0, "object_id": "6"},
      "7": {"semantics": "table", "raw_label": "coffee table",  "center": [0.0, -0.5, 0.3], "volume": 0.5, "object_id": "7"},
      "8": {"semantics": "plant", "raw_label": "plant",         "center": [0.0, -0.5, 0.8], "volume": 0.1, "object_id": "8"}
    },
    "region_id": 1
  }
}

Expected assistant response:
{"hypotheses": [
  {"utterance": "there is a mouse on the desk",  "referred_object_id": ["0"], "region_id": 0, "confidence": 0.8},
  {"utterance": "there is a mouse on the table", "referred_object_id": ["7"], "region_id": 1, "confidence": 0.3}
]}
\end{lstlisting}

\section{Foundation Model Baseline Prompts}
\label{app:baseline_prompts}

Prompts used for the three foundation-model baselines from Section~\ref{sec:experiments}.

\paragraph{Scaffolded-LLM.} The system prompt for the per-hypothesis LLM Gaussian regressor:

\begin{lstlisting}[style=prompt]
You are a spatial reasoning module that predicts the 3D location of an object described by a natural-language utterance.

Given:
- A scene graph: objects (with world-frame centres) grouped by region.
- An utterance introducing a new (target) object.
- A grounding: the region and anchor object(s) the utterance refers to (assumed correct).

Output a 3D Gaussian distribution N(mu, Sigma) over the TARGET object's world-frame position.

Format strictly as JSON:
{
  "mu":    [x, y, z],
  "sigma": [[s11,s12,s13],[s12,s22,s23],[s13,s23,s33]],
  "explanation": "..."
}

Guidance for Sigma:
- Use tighter sigma (e.g. 0.05-0.2 m) for precise relations like "on", "in".
- Use looser sigma (e.g. 0.5-2.0 m) for vague relations like "near", "around".
- Sigma must be symmetric and positive semi-definite.
\end{lstlisting}

\paragraph{LLM-E2E.} The system prompt for the end-to-end LLM that emits the full mixture in one call:

\begin{lstlisting}[style=prompt]
You are a spatial reasoning module that predicts the 3D location of a target object described by a natural-language utterance, *without* being told which anchor objects the utterance refers to.

Given:
- A scene graph: objects (with world-frame centres) grouped by region.
- An utterance introducing a new target object.

Output a Gaussian mixture model over the new target object's world-frame position - one component per plausible grounding of the utterance.

Format strictly as JSON:
{
  "components": [
    {
      "mu":    [x, y, z],
      "sigma": [[...3x3...]],
      "weight": 0.5,
      "explanation": "..."
    },
    ...
  ]
}

- Weights must be non-negative and sum to 1.
- Return at least 1 and at most 5 components (one per plausible anchor).
- Sigma must be symmetric and positive semi-definite.
\end{lstlisting}

\begin{figure}[!htb]
    \centering
    \includegraphics[width=0.4\textwidth,  angle=-90]{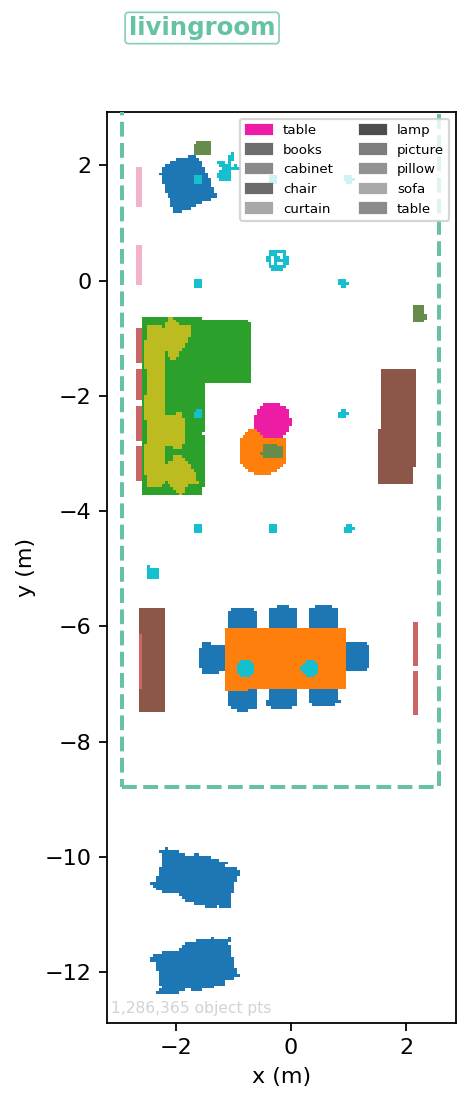}
    \caption{Example BEV prompt: point cloud coloured by semantics, anchor object in magenta.}
    \label{fig:bev_example}
\end{figure}

\paragraph{Scaffolded-VLM.} The VLM consumes a top-down BEV of the scene's semantic point cloud, filtered to the floor and coloured by object class, with the anchor object highlighted in a distinguishing colour. Figure~\ref{fig:bev_example} shows an example BEV; the accompanying utterance is \textit{``There is a bag near the small table.''} The system prompt:

\begin{lstlisting}[style=prompt,caption={\textbf{Scaffolded-VLM} system prompt.}]
You are a spatial reasoning module that predicts the 3D location of a target object described by a natural-language utterance.

You are given:
- A Bird's-Eye-View (BEV) image of the relevant region. Each non-structural object is drawn as a rectangle labeled with its semantic class; the magenta-filled rectangles are the anchor objects the utterance references.
- A natural-language utterance introducing the target object (not yet on the map).
- A text description of the anchors and region.

Output a 3D Gaussian N(mu, sigma) over the target object's world-frame position. The BEV image x/y axes are world-frame metres; the z (height) must be inferred from context (floor, table-top, etc.).

Format strictly as JSON:
{
  "mu":    [x, y, z],
  "sigma": [[s11,s12,s13],[s12,s22,s23],[s13,s23,s33]],
  "explanation": "..."
}
\end{lstlisting}

\paragraph{Implementation details shared across baselines.} All baseline responses are parsed with the same JSON-extraction routine, with the predicted covariance projected to the positive-semi-definite cone via an eigenvalue clip at $10^{-2}\,\mathrm{m}^2$ before Cholesky factorization. 

\section{Language Sensor Model Implementation Details}
\label{app:lsm_impl}
\label{app:lsm_implementation}

This section pins down the architectural sizes and training objective behind the \lsm{} sketched in Section~\ref{sec:lsm}. All dimensions are listed in Table~\ref{tab:lsm_arch}. The text and visual backbones are kept frozen; all remaining modules are trained jointly.

\begin{table}[!htb]
    \centering
    \small
    \caption{\textbf{\lsm{} architecture.} All hidden dimensions and layer counts. Region scale/shift $(s,t) \in \R^3 \times \R^3$ are precomputed from the anchor's local region and map the head's predictions back to world coordinates.}
    \begin{tabular}{ll}
        \toprule
        Component & Setting \\
        \midrule
        Hidden dimension          & $256$ \\
        Text encoder              & BERT-base-uncased (frozen), $768 \!\to\! 256$, max $64$ tokens \\
        Object encoder            & CLIP ViT-B/32 (frozen), $512 \!\to\! 256$, $+$ role $+$ modality embeddings \\
        Max objects per scene     & $150$ \\
        Query token               & learnable $[Q_{\mathrm{target}}]$, $256$-d, prepended for pooling \\
        Spatial backbone          & $3$ layers, $8$ heads, FFN $1024$, pre-norm, GELU \\
        Pairwise spatial feature  & $12$-d anchor-centric (relative centroid, size, distance) \\
        Spatial bias              & MLP $12 \!\to\! 64 \!\to\! 8$ (one bias per head), FiLM-conditioned on text \\
        Fusion backbone           & $3$ layers, $8$ heads, FFN $1024$, pre-norm, GELU \\
        FiLM head conditioning    & MLP, hidden $64$, modulates pooled context by region $(s,t)$ \\
        Output head               & $256 \!\to\! 1024 \!\to\! 512 \!\to\! 9$, GELU, dropout $0.15$ \\
        Gaussian parameterisation & $\gmean \in \R^3$ (\texttt{tanh}); $L \in \R^{3 \times 3}$ lower-tri., softplus diag., $\sigma$-floor $0.05$ \\
        World-frame Gaussian      & $\gmean_w = s \odot \gmean + t,\ \ L_w = \mathrm{diag}(s)\,L,\ \ \gcov = L_w L_w^{\!\top}$ \\
        \bottomrule
    \end{tabular}
    \label{tab:lsm_arch}
\end{table}

\paragraph{Loss.} Given a ground-truth target centroid $y \in \R^3$ and predicted Gaussian $\Norm(\gmean, \gcov)$, the model is trained with the additive objective
\[
\mathcal{L} \;=\; \tfrac{1}{2}\!\left( \log\!\lvert \gcov \rvert + (y - \gmean)^{\!\top} \gcov^{-1} (y - \gmean) + 3 \log 2\pi \right) \;+\; \lambda_1\,\lVert \gmean - y \rVert_1 ,
\]
with $\lambda_1 = 1.0$. The first term is the standard Gaussian negative log-likelihood; the second is a direct $\ell_1$ regression that anchors the mean independently of $\gcov$.
\cameraready{\paragraph{Inference latency.}
In the latency evaluation, the hypothesis proposer and all three
foundation-model baselines use GPT-5.2.
Table~\ref{tab:language_prediction_latency} reports average end-to-end
language-prediction time. For \lsm{}, this includes the LLM hypothesis
proposal and transformer forward pass. The scaffolded baselines
require additional per-hypothesis queries. Language predictions enter
\vlmap{} asynchronously, while visual integration runs at sensor
frame rate.}

\begin{table}[!htb]
    \centering
    \small
    \caption{Average end-to-end language-prediction latency.}
    \label{tab:language_prediction_latency}
    \begin{tabular}{lc}
        \toprule
        Method & Latency (s) $\downarrow$ \\
        \midrule
        LLM-E2E        & 8.38  \\
        Scaffolded-LLM & 8.61  \\
        Scaffolded-VLM & 15.82 \\
        \lsm{} (ours)  & 2.32  \\
        \bottomrule
    \end{tabular}
\end{table}
\newpage
\section{Per-Relation Grounding Breakdown}
\label{app:per_relation}

\begin{figure}[!htb]
    \centering
    \includegraphics[width=0.7\textwidth]{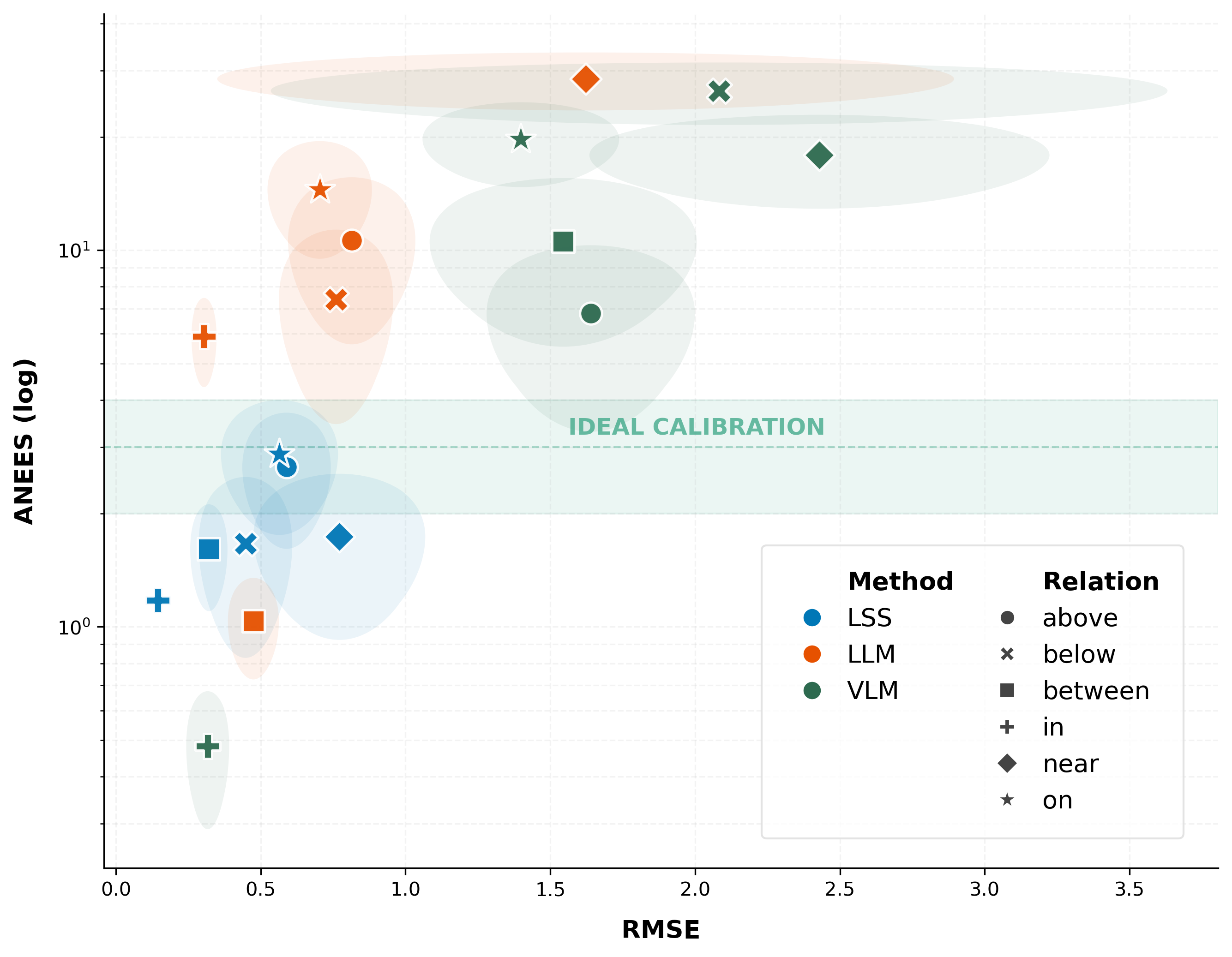}
    \caption{\textbf{Per-relation grounding accuracy versus calibration on unambiguous utterances.} \lsm{} is the only method that clusters near the calibrated optimum $(\text{ANEES}=3,\,\text{RMSE}=0)$ across all six spatial relations, while the foundation-model baselines collapse on extent-heavy relations such as \emph{near}, \emph{between}, and \emph{above}, where the language most underdetermines the valid region. Each method contributes six points (one per relation: \emph{above, below, between, in, near, on}) with $\pm$std ellipses.}
    \label{fig:unambiguous_performance}
\end{figure}

Section~\ref{subsec:grounding} references the per-relation breakdown that exposes \emph{where} foundation-model groundings fail. Figure~\ref{fig:unambiguous_performance} shows the unambiguous-utterance result disaggregated by spatial relation.

\end{document}